\documentclass{article}
\usepackage{PRIMEarxiv}
\usepackage[utf8]{inputenc}
\usepackage[T1]{fontenc}

\usepackage{amsmath,amsfonts,bm}

\def\eqref#1{equation~\ref{#1}}

\def\1{\bm{1}}

\DeclareMathAlphabet{\mathsfit}{\encodingdefault}{\sfdefault}{m}{sl}
\SetMathAlphabet{\mathsfit}{bold}{\encodingdefault}{\sfdefault}{bx}{n}

\usepackage{amsmath}
\usepackage{amssymb}
\usepackage{array}
\usepackage{mathtools}
\usepackage{booktabs}
\usepackage{graphicx}
\usepackage[percent]{overpic}
\usepackage{microtype}
\usepackage{placeins}
\usepackage[round,authoryear]{natbib}
\usepackage{hyperref}
\usepackage{url}
\hypersetup{hidelinks}
\newcommand{\method}{RemTraceNet}
\newcommand{\tprone}{TPR@1\%FPR}
\newcommand{\tprfive}{TPR@5\%FPR}

\title{\method: Few-Shot Forensic Detection of Invisible Watermark Attacks}

\author{
\parbox{0.96\textwidth}{\centering
Jidong Yang\textsuperscript{1}, Huaike Yu\textsuperscript{1},
Qi Li\textsuperscript{1}\thanks{Corresponding author: Qi Li
(e-mail: qluliqi@163.com).}, Yuantian Miao\textsuperscript{2},
Wei Zong\textsuperscript{3} \\
Yang-Wai Chow\textsuperscript{3}, Willy Susilo\textsuperscript{3},
Chunpeng Wang\textsuperscript{1}, Suo Gao\textsuperscript{4} \\
\normalfont\small \textsuperscript{1}Key Laboratory of Computing Power Network and Information Security,
Ministry of Education; Shandong Computer Science Center; \\
\normalfont\small Shandong Provincial Key Laboratory of Industrial Network and Information System Security;
Shandong Fundamental Research Center for Computer Science; \\
\normalfont\small Qilu University of Technology (Shandong Academy of Sciences), Jinan 250353, China \\
\normalfont\small \texttt{jidong\_yang\_paper@163.com; huaikeyu@gmail.com; mpeng1122@163.com} \\
\normalfont\small \textsuperscript{2}Department of Computer Science,
City University of Hong Kong (Dongguan), Dongguan, Guangdong 518057, China \\
\normalfont\small \texttt{yuantian.miao@cityu-dg.edu.cn} \\
\normalfont\small \textsuperscript{3}Institute of Cybersecurity and Cryptology (iC2),
University of Wollongong, Australia \\
\normalfont\small \texttt{wzong@uow.edu.au; caseyc@uow.edu.au; wsusilo@uow.edu.au} \\
\normalfont\small \textsuperscript{4}School of Information Science and Engineering,
Dalian Polytechnic University, Dalian 116034, China \\
\normalfont\small \texttt{gaosuodlpu@163.com}
}
}
\begin{document}

\maketitle

\begin{abstract}
Removing an invisible watermark and concealing the forensic evidence are
distinct objectives: successfully disrupting the embedded watermark does not
imply that the removal process is forensically undetectable. When verification
fails, removal traces can provide complementary evidence for provenance and
ownership verification, whereas their absence leaves the cause of the failure
ambiguous. Existing methods are typically evaluated by watermark suppression
and perceptual quality, while forensic stealth is rarely considered. We
therefore study watermark-attack-specific few-shot forensics: for each known
pipeline, a specialist can separate its outputs from paired clean and
unattacked watermarked controls. Separate Attack-vs-Clean and
Attack-vs-Watermarked evaluations prevent watermark-presence shortcuts.
Image-aligned and prompt-matched controls are used for post-hoc and
generator-integrated schemes, respectively. In this work, we introduce
\method, which fuses constrained residuals, local relations, FFT/Haar
statistics, and block-DCT evidence at native resolution. Across 23 removal
pipelines and 10 watermark configurations, we evaluate native
$256\!\times\!256$ and $512\!\times\!512$ inputs. With 100 attacked training
images per pipeline, the three-seed \tprone{}, macro-averaged over attacks and
watermark configurations, ranges from 82.75\% to 88.17\% across resolutions and
control types. Under the condition of same labels and protocol, \method{}
outperforms retrained SRNet, ZhuNet, and SiaStegNet baselines by
10.80--15.68 percentage points. Extensive experimental results show that
erasing a watermark and erasing evidence of its removal are distinct
challenges, and that removal traces remain learnable under limited supervision.
\end{abstract}


\section{Introduction}
\label{sec:introduction}

Invisible image watermarking embeds machine-readable signals while limiting
visible distortion. Neural methods now span post-hoc pixel-space encoders,
latent representations, and generator-integrated signals
\citep{zhu2018hidden,tancik2020stegastamp,fernandez2022watermarking,
fernandez2023stablesignature,wen2023treering,yang2024gaussianshading}.
Combined with a detector, a key, and a registry, they can support provenance
tracking and ownership verification. Their practical value, however, depends
not only on whether the embedded watermark remains recoverable after
processing, but also how a failed verification is interpreted. Imperceptible
removal does not imply forensic stealth: a removal pipeline may disrupt
watermark recovery while leaving processing traces. Such traces may offer
complementary evidence for future provenance and ownership analysis when direct
watermark verification fails.

Robustness evaluation typically applies distortions or removal pipelines and
then queries the target detector or decoder. The attack surface includes signal
and geometric transformations, detector-aware optimization, learned
reconstruction, diffusion-based regeneration, latent inversion, and forgery
\citep{petitcolas1998attacks,jiang2023wevade,lukas2024leveraging,
zhao2024provably,liu2025ctrlregen,qiu2025nfpa,wmcopier2025}. These protocols
measure watermark detectability or payload recovery, not whether the processed
output remains distinguishable from a clean or unattacked-watermarked control.
Consequently, an attack may make the payload undecodable while leaving
residual, spectral, reconstruction, or spatial inconsistencies. Conversely, a
failed verification may also result from benign post-processing during
transmission.

This motivates watermark-attack-specific few-shot forensics. Our objective is
to identify whether an image was processed by a specified removal pipeline,
rather than to detect watermark presence or determine whether the payload has
been successfully removed. Existing watermark detectors recognize or decode
payload evidence rather than this processing history. In this paper, for each
known removal pipeline, we train a separate binary specialist that classifies
attacked outputs as positive and paired clean and unattacked-watermarked
controls as negative. ``Attack-specific'' denotes a checkpoint and label
construction for a known removal pipeline.

Visual similarity does not imply forensic indistinguishability. For each
attacked output, we construct paired clean and unattacked-watermarked controls:
post-hoc pairs share a source image, whereas generator-integrated pairs share a
prompt. The specialist is trained jointly on all three roles, with both controls
negative. Attack-vs-Clean tests separation from the paired unprocessed control
under Section~\ref{sec:problem}'s alignment regime; Attack-vs-Watermarked tests
additional processing rather than watermark presence. Separate reporting
exposes false positives that pooled controls could hide, while We keep content
identities disjoint between fitting and checkpoint-selection splits while
preserving the prescribed pairing.

The specialist must distinguish heterogeneous processing evidence from image
content. To address this challenge, we introduce \method, a native-resolution
forensic architecture that constructs four complementary evidence views:
global residuals, local relations, FFT/Haar statistics, and block-DCT evidence.
It combines a global prediction with Top-$k$ aggregation of dense local
evidence.

We conduct a controlled empirical study across 23 removal pipelines and ten
watermark configurations. Each specialist is trained on five watermark
configurations and evaluated on ten configurations, including five unseen
during fitting. The evaluation uses native 256x256 and 512x512 inputs, three
random seeds, and attacked training sizes ranging from $n = 10$ to 200. The
split design keeps fitting and checkpoint-selection content identities
disjoint across watermark configurations. At $n = 100$, the three-seed mean
\tprone{} is 87.35\% and 88.17\% at 256x256 for clean and
unattacked-watermarked negatives, respectively, and 82.75\% and 82.88\% at
512x512. Under a seed-matched comparison with the same labels and evaluation
protocol, \method{} exceeds the strongest retrained steganalysis architecture
by 10.80--15.68\% across these four settings. After macro-averaging, the two
negative-control settings differ by only 0.82\% at resolution 256 and 0.13\% at
resolution 512. For unseen watermarks, \tprone{} is lower than for seen
watermarks in all four settings.

Our contributions are as follows:
\begin{enumerate}
    \item We formulate watermark-attack-specific few-shot forensics as a
    processing-history detection task with paired Attack-vs-Clean and
    Attack-vs-Watermarked controls, distinguishing image-aligned post-hoc pairs
    from prompt-matched generator-integrated pairs.
    \item We introduce \method, a native-resolution multi-view detector that
    combines residual, relational, spectral, and block-DCT evidence with dense
    local multiple-instance decisions.
    \item We report a controlled empirical study across 23 pipelines and 10
    watermark configurations. The evaluation covers two native resolutions,
    three seeds, training sizes from 10 to 200, task-adapted steganalysis
    baselines, and component ablations.
    \item We show that imperceptible watermark-removal attacks can leave
    learnable forensic traces. By detecting these traces, \method{} highlights
    the potential of removal-specific forensics to support future copyright and
    provenance when direct verification fails.
\end{enumerate}

\begin{figure}[t]
  \centering
  \includegraphics[width=\textwidth]{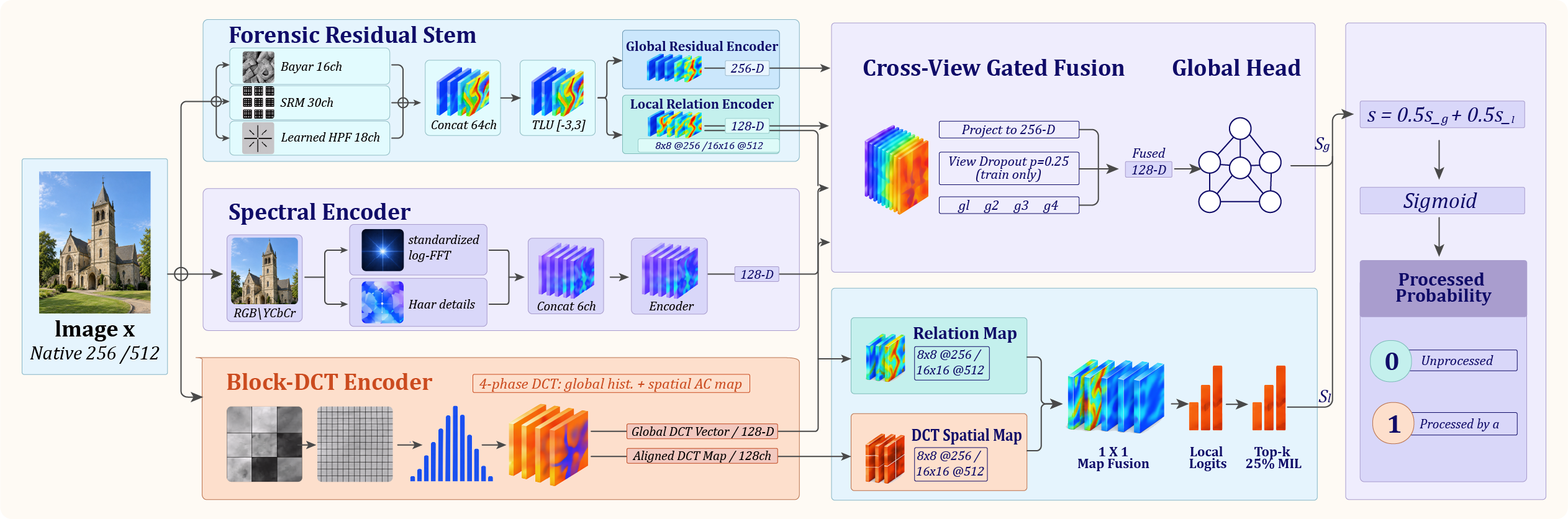}
  \caption{\method{} overview. One specialist is selected for a known
  pipeline, while the image input contains no pipeline or watermark identity.
  Four global views are fused by learned gates; dense relation and DCT maps
  form a local Top-$k$ MIL decision. Equal weighting of the global and local
  logits produces the probability that the image was processed.}
  \label{fig:overview}
\end{figure}

\section{Problem Formulation}
\label{sec:problem}

Let $q$ denote the pairing unit and let $(x_q^0,x_{q,w})$ denote its clean and
unattacked watermarked controls. For post-hoc watermarking, $q=c$ is a source
identity, $x_q^0=x_c$, and $x_{q,w}=W_w(x_c)$. This forms an image-aligned
pair. For generator-integrated watermarking, $q=p$ is a prompt,
$x_q^0=G_0(p,\xi_0)$, and $x_{q,w}=G_w(p,\xi_w)$. This forms a prompt-matched
pair of distinct realizations. Both regimes retain an explicit clean/watermarked
pair through $q$; only the former assumes pixel correspondence.

For pipeline $a$, $x^{+}_{q,w,a}$ is the processed watermarked member; ordinarily
$x^{+}_{q,w,a}=K_a(x_{q,w})$, while forgery may also use donor or reference
inputs. One specialist $f_a(x)\in\mathbb{R}$ is trained on
$\{x_q^0,x_{q,w},x^{+}_{q,w,a}\}$, with the processed image labeled one and both
controls zero. The attacked class is balanced against the combined negatives;
no separate clean detector is trained. The specialist checkpoint corresponds to
$a$, but the input contains no attack or watermark identifier.

The same specialist checkpoint is reported in two binary slices:
\begin{align}
\mathcal{T}_{\mathrm{clean}}(a,w) &:
  \{x^{+}_{q,w,a}\}_{y=1}\ \text{vs.}\ \{x_q^0\}_{y=0}, \\
\mathcal{T}_{\mathrm{wm}}(a,w) &:
  \{x^{+}_{q,w,a}\}_{y=1}\ \text{vs.}\ \{x_{q,w}\}_{y=0}.
\end{align}
The first task compares against paired unprocessed content. For image-aligned
pairs it tests the combined embedding and attack history against the exact source;
for prompt-matched pairs it also contains generation variability. The second
prevents watermark presence from becoming sufficient positive evidence. Each
watermark is reported before macro aggregation. The detector receives no key,
encoder, decoder, target detector, or attack parameters.

Operationally, this is attack-specific forensics because the specified
pipeline produces the positive from the watermarked member and selects its
specialist checkpoint. Table~\ref{tab:notation} summarizes the notation used
in the formulation and model.

\begin{table}[!ht]
\caption{Symbols used in the problem formulation and method.}
\label{tab:notation}
\centering
\footnotesize
\setlength{\tabcolsep}{3pt}
\renewcommand{\arraystretch}{0.92}
\begin{tabular}{@{}>{\raggedright\arraybackslash}p{0.12\linewidth}>{\raggedright\arraybackslash}p{0.35\linewidth}>{\raggedright\arraybackslash}p{0.12\linewidth}>{\raggedright\arraybackslash}p{0.35\linewidth}@{}}
\toprule
Symbol & Meaning & Symbol & Meaning \\
\midrule
$q$ & Pairing unit & $c$, $p$ & Source identity and prompt \\
$w$, $a$ & Watermark configuration and specified pipeline &
$W_w$, $K_a$ & Watermark embedding and pipeline operators \\
$G_0$, $G_w$ & Clean and watermarked generators &
$\xi_0$, $\xi_w$ & Generation random variables \\
$x_c$, $x_q^0$ & Source image and paired clean control &
$x_{q,w}$, $x^+_{q,w,a}$ & Unattacked and processed watermarked images \\
$y$ & Binary label: processed $1$, control $0$ &
$f_a(x)=s$ & Logit of the specialist for pipeline $a$ \\
$\mathcal{T}_{\mathrm{clean}}$, $\mathcal{T}_{\mathrm{wm}}$ &
Clean- and watermarked-control evaluation slices &
$n$ & Number of complete training triplets \\
$r_i$, $\mathcal{N}(i)$ & Region descriptor and its neighborhood &
$|\cdot|$, $\odot$, $\mathcal{F}$ & Elementwise absolute value, product, and 2-D Fourier transform \\
$z_j$, $u_j$ & $j$th global view and its projection &
$G$, $H$ & Gate map and fused-representation map \\
$g$, $g_j$, $z$ & Gate vector, view weight, and fused representation &
$s_g$ & Global logit \\
$\ell_m$, $M$ & $m$th local logit and number of local instances &
$k$ & Number of selected local logits, $\lceil0.25M\rceil$ \\
$\operatorname{TopK}(\ell,k)$ & Indices of the $k$ largest local logits &
$s_l$, $s$ & Local and final logits \\
$\mathcal{L}_{\mathrm{BCE}}$, $\mathcal{L}_{\mathrm{rank}}$, $\mathcal{L}_{\mathrm{pair}}$ &
Classification, low-FPR ranking, and paired losses &
$\mathcal{L}_{\mathrm{branch}}$, $\mathcal{L}_{\mathrm{gate}}$, $\mathcal{L}_{\mathrm{HPF}}$ &
Branch, gate-balance, and high-pass penalties \\
AUROC, AUPRC & Areas under the ROC and precision--recall curves &
TPR, FPR & True- and false-positive rates \\
\bottomrule
\end{tabular}
\end{table}
\FloatBarrier

\section{Method}
\label{sec:method}

\subsection{Multi-view forensic representation}

Figure~\ref{fig:overview} illustrates the overall architecture of  \method{}. Under the paired forensic task, the
model must distinguish heterogeneous processing evidence from image content
without treating watermark presence alone as positive. We hypothesize that no
single statistic is sufficient across the pipeline set. \method{} therefore
extracts four vectors: a 256-dimensional
global residual vector and 128-dimensional relation, spectral, and DCT
vectors. The relation and DCT paths additionally retain spatial maps. The
full model contains 4.80 million trainable parameters.

\paragraph{Forensic residual stem and global encoder.}
The inputs are rescaled from $[0,1]$ to $[0,255]$ after residual filtering.
The learned $5!\times!5$ Bayar filter fixes its center coefficient to $-1$ and projects the remaining coefficients to sum to one. Consequently, the complete filter has a zero coefficient sum, which suppresses its DC response \citep{bayar2018constrained}. A fixed 30-filter SRM bank and an
18-channel learned $3\!\times\!3$ high-pass path complement its 16 output
channels. Its penalty sums, over all input-output channel pairs, the absolute
value of each two-dimensional kernel's coefficient sum. We
concatenate all 64 channels and apply a truncated linear unit at $[-3,3]$.
Group-normalized residual blocks downsample this tensor by a factor of 16;
global mean and standard deviation pooling produce the 256-dimensional view.

\paragraph{Local relation view.}
The residual tensor is shallowly encoded and then adaptively pooled into regions with a physical extent of 32 pixels. This produces an $8\!\times\!8$ grid at
resolution 256 and a $16\!\times\!16$ grid at resolution 512. For each region
$r_i$ and its $3\!\times\!3$ neighborhood $\mathcal{N}(i)$, the edge model
uses $|r_j-r_i|$ and $r_j\odot r_i$ to predict attention weights over
$j\in\mathcal{N}(i)$. A residual update yields a 128-channel relation map;
mean and standard deviation pooling form its global vector.

\paragraph{Spectral view.}
The RGB image is converted to the YCbCr color space. We compute a centered
$\log(1+|\mathcal{F}(x)|)$ spectrum for each channel and standardize it per
image and channel. Three single-level Haar detail bands (LH, HL, and HH) from
the luminance channel are aligned with the spectra. A residual encoder then
maps the six channels to a 128-dimensional vector. The FFT path provides a
global frequency summary, while the Haar bands provide directional
high-frequency coefficients.

\paragraph{Block-DCT view.}
The luminance channel is transformed uesing orthonormal $8\!\times\!8$ DCT
blocks at four offsets $(0,0)$, $(0,4)$, $(4,0)$, and $(4,4)$. For each of the 63
AC frequencies, differentiable histograms combine unit-spaced bins over
$[-16,16]$, stride-four bins over $[-128,128]$, and two overflow bins. After
normalization and a square-root transform, a convolution over the restored
$8\!\times\!8$ frequency plane yields a 128-dimensional global vector.
Signed log-magnitude AC maps are encoded separately, aligned across phase
offsets, and averaged to retain block-level spatial evidence. The supplementary
material provides a consolidated diagram of the four evidence-view construction
paths.

\subsection{Global and local decisions}

Let $z_j$ be the $j$th global view. A view-specific linear map projects every
$z_j$ to $u_j\in\mathbb{R}^{256}$. The learned global path computes
\begin{equation}
g=\operatorname{softmax}\!\left(G[u_1;\ldots;u_4]\right),\qquad
z=H\!\left(\sum_{j=1}^{4}g_j u_j\right),
\end{equation}
where $H$ produces a 128-dimensional representation. During training, each
view is independently dropped with probability 0.25 while retaining at least
one view, and the surviving gate weights are renormalized. A two-layer head
maps $z$ to the global logit $s_g$.

For the local path, $1\!\times\!1$ projections align the relation and DCT maps
to 128 channels and the relation grid. A convolution produces local logits
$\{\ell_m\}_{m=1}^{M}$. With $k=\lceil0.25M\rceil$, the local logit is
\begin{equation}
s_l=\frac{1}{k}\sum_{\mathclap{m\in\operatorname{TopK}(\ell,k)}}\!\ell_m,
\qquad s=0.5s_g+0.5s_l.
\end{equation}
The sigmoid of $s$ is the processed-image probability. This path retains
spatially indexed logits until the final pooling step. Section~\ref{sec:ablations}
tests the components whose removal consistently reduces the primary metric.

\subsection{Role-balanced low-FPR objective}

The training objective follows the two-negative task definition rather than collapsing clean and unattacked-watermarked controls into a single undifferentiated class.
Each training set contains $n$ complete triplets, hence $n$ attacked images
and $n$ images from each negative role. A role-balanced binary cross-entropy
assigns effective mass 0.50 to attacked examples and 0.25 to each negative
role. These masses weight all observed samples; they do not subsample either
negative role. A ranking term compares positives separately against the
hardest 20\% of clean and watermarked negatives, using a soft margin of 0.5;
the two losses are averaged rather than pooling their tails. For each pairing
unit $q$, the paired term pulls the normalized clean and watermarked
features together and pushes attacked features at least one unit from their
negative center. The term uses the available alignment level and does not
assume pixel correspondence for prompt-matched generator-integrated pairs.

The full objective is
\begin{equation}
\mathcal{L}=\mathcal{L}_{\mathrm{BCE}}
+0.5\mathcal{L}_{\mathrm{rank}}
+0.2\mathcal{L}_{\mathrm{pair}}
+0.2\mathcal{L}_{\mathrm{branch}}
+0.01\mathcal{L}_{\mathrm{gate}}
+0.001\mathcal{L}_{\mathrm{HPF}}.
\end{equation}
The branch term applies the role-balanced BCE to each global view. The gate
term penalizes deviation of the batch-mean gate from uniform use, and the HPF
term enforces the learned high-pass prior. These terms encourage separation
under the available paired controls.
The supplementary material provides the complete definitions and implementation
details for all objective terms.

\section{Experiments}
\label{sec:experiments}

\subsection{Data and strict protocol}

All images are drawn from the MS-COCO dataset \citep{lin2014coco}. We use five watermarking configurations for fitting and checkpoint selection:
DwtDctSvd \citep{kang2018novel}, Stable Signature
\citep{fernandez2023stablesignature}, Gaussian Shading
\citep{yang2024gaussianshading}, SSL Watermarking
\citep{fernandez2022watermarking}, and EditGuard \citep{zhang2024editguard}.
The unseen group contains five additional configurations: DwtDct \citep{kang2018novel}, HiDDeN
\citep{zhu2018hidden}, StegaStamp \citep{tancik2020stegastamp}, Tree-Ring
\citep{wen2023treering}, and RivaGAN \citep{zhang2019rivagan} configurations.
These five unseen configurations contribute no training or validation images.
Watermarking configuration identities are used only for sampling and reporting and are not provided to the model. Post-hoc pairs share the same source image; Stable Signature, Gaussian
Shading, and Tree-Ring pairs share a prompt but contain distinct realizations.
Both use one manifest pairing identifier. Training and validation splits
enforce stem disjointness across watermarks. The formal test set with ten
watermarks excludes validation stems but
does not deduplicate stems across watermark configurations; metrics are first
computed per watermark.

Each experimental run evaluates a single attack pipeline. The 23 positive pipelines comprise 12 signal
or geometric transformations, five learned VAE compression/reconstruction
models \citep{balle2018variational,minnen2018joint,cheng2020learned,
begaint2020compressai},
three diffusion regeneration attacks \citep{zhao2024provably,
liu2025ctrlregen,sadre2025}, NFPA \citep{qiu2025nfpa}, BM3D
\citep{dabov2007image}, and WMForger \citep{soucek2025wmforger}. For each pipeline, the positive label indicates that the specified operation produced
the image; it does not indicate that the operation necessarily removed a
watermark.

Training sizes are $n\in\{10,20,\ldots,100,150,200\}$ complete triplets per
attack, distributed evenly over five training watermarks. Each triplet contains
one attacked image, one clean control, and one unattacked watermarked control.
Thus, $n=200$ means 200 images in each role. After the single 1,000-update
training epoch, the final EMA weights are evaluated once on the fixed validation
set and stored as the reported checkpoint. Formal evaluation then compares, for
each watermark and resolution, 100 attacked images with 100 clean images and,
separately, with 100 watermarked images.
The supplement provides a schematic of the data construction and evaluated
generalization axes.

\subsection{Optimization and metrics}

We train each model for exactly 1,000 optimizer updates with eight content groups per
update. We use AdamW \citep{loshchilov2019decoupled} with an initial learning
rate of $4\times10^{-4}$, weight decay $10^{-4}$, and 50 warm-up steps. The
schedule uses cosine decay to $10^{-6}$, mixed precision, and exponential
moving average decay 0.999. The main 23-pipeline scaling sweep uses seeds 42,
43, and 44. All models are trained on native $256\!\times\!256$ inputs. The same checkpoint is then evaluated on native $256\!\times\!256$ and $512\!\times\!512$ inputs without external resizing.

We report AUROC, AUPRC, \tprfive{}, and \tprone{}, with \tprone{} serving as the primary metric.
Both operating-point metrics use test scores. For each slice defined by attack,
watermark, resolution, and negative role, \tprone{} is read from the empirical
test ROC. It is the largest attainable TPR at an FPR not exceeding 1\%.
The resulting thresholds are slice-specific descriptive operating points,
rather than a single validation-calibrated deployment threshold. Metrics are
computed per slice and seed, macro-averaged over attacks and watermarks, and
then summarized across three seeds. Each slice has 100 negatives, so zero, one,
and two false positives give empirical FPRs of 0\%, 1\%, and 2\%. One
high-scoring negative can move the threshold. Tied scores can skip a 1\%
increment.

\subsection{Task-adapted steganalysis baselines}

To compare architectures under the same forensic task, SRNet, ZhuNet, and SiaStegNet are retrained from scratch for the same binary
labels, manifests, role balance, 1,000-update budget, optimizer schedule, EMA,
and native-resolution tests. Each baseline uses role-balanced two-class
cross-entropy; SiaStegNet's optional contrastive term is disabled. \method{}
retains its full objective, so this comparison evaluates complete task-adapted
systems rather than isolating architecture under a shared loss. Model-specific
input conventions are preserved; the supplement provides implementation
details. We perform no model-specific hyperparameter search. The baseline comparison uses seed 42 for all architectures, so it is a seed-matched architectural comparison, not a three-seed significance analysis.

\subsection{Main results and generalization analysis}

Table~\ref{tab:main} summarizes empirical test-ROC performance with $n=100$
complete training triplets per attack.
The \tprone{} values averaged over all watermarks differ between the two negative roles by
0.82\% at resolution 256 and 0.13\% at resolution 512. At this aggregation
level, performance is comparable under the two controls.
For unseen watermarks, \tprone{} trails performance on seen watermarks by 7.84\% and
7.71\% at resolution 256. At resolution 512, the gap is 0.43\% against clean
negatives and 5.49\% against watermarked negatives. The gap for unseen watermarks
therefore ranges from 0.43--7.84\% across the four settings. Performance at
resolution 512 is also lower than at 256, so native-resolution evaluation
should not be read as resolution invariance.

\begin{table}[t]
\caption{Main test-ROC results with $n=100$ training triplets per attack, reported as mean $\pm$ standard deviation (\%) over three seeds, with ``All'' averaging five seen and five unseen watermarks.}

\label{tab:main}
\centering
\scriptsize
\begin{tabular}{@{}cccccc@{}}
\toprule
Res. & Negative & \shortstack{AUROC\\(All)} & \shortstack{\tprone{}\\(All)} &
\shortstack{\tprone{}\\(Seen)} & \shortstack{\tprone{}\\(Unseen)} \\
\midrule
256 & Clean       & $98.21\pm0.09$ & $87.35\pm0.90$ & $91.28\pm0.98$ & $83.43\pm0.90$ \\
256 & Watermarked & $98.01\pm0.06$ & $88.17\pm0.36$ & $92.03\pm0.88$ & $84.31\pm0.56$ \\
512 & Clean       & $96.44\pm0.62$ & $82.75\pm1.19$ & $82.97\pm1.80$ & $82.53\pm0.58$ \\
512 & Watermarked & $96.68\pm0.69$ & $82.88\pm1.74$ & $85.62\pm2.40$ & $80.13\pm1.44$ \\
\bottomrule
\end{tabular}
\end{table}

\begin{figure}[!t]
  \centering
  \includegraphics[width=\textwidth,height=0.70\textheight,keepaspectratio]{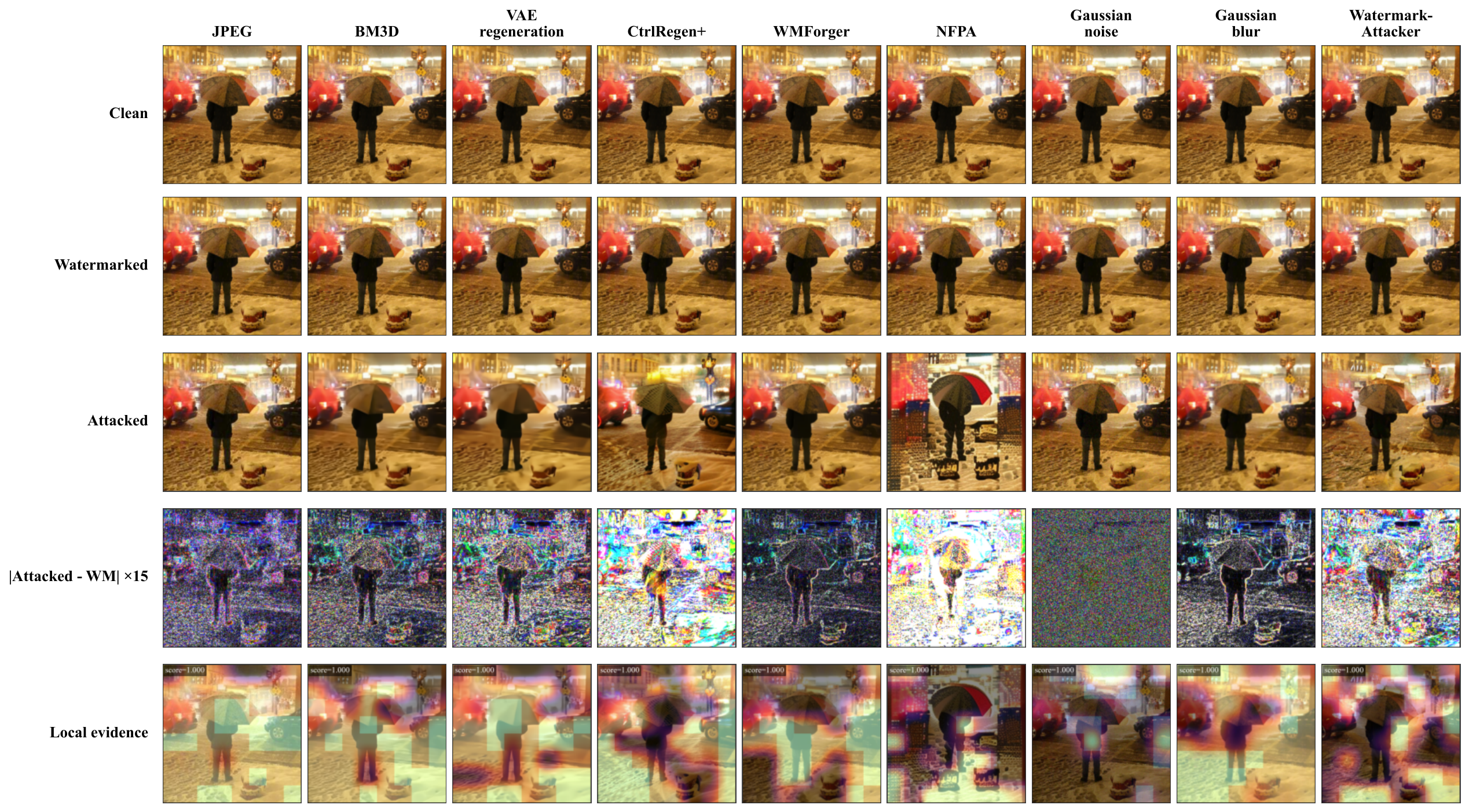}
  \caption{Qualitative forensic traces across nine attacks,
  showing the clean, watermarked, and attacked images, the amplified
  attacked--watermarked residual ($\times15$), and the native local-decision
  response for the same input.}
  \label{fig:trace-comparison}
\end{figure}

\paragraph{Qualitative analysis.}
Figure~\ref{fig:trace-comparison} shows a paired DwtDctSvd example for nine
attacks at seed 42, native resolution 256, and $n=100$. The
visible changes and amplified attacked--watermarked residuals differ across
noise, filtering, regeneration, inversion, and forgery operations. The final
row is the native local-decision response before Top-$k$ pooling. It is a
diagnostic of the local decision path, not a pixelwise attribution map, and the
example does not establish strict alignment with the residual or independence
from image semantics. Additional examples appear in the supplement.

\FloatBarrier

The sweep over training size evaluates data efficiency for a specialist trained on
a known pipeline under a fixed 1,000-update budget. The supplement reports
$n=10$--200 aggregate curves, seed-42 operation-level curves at resolution 256,
seed-specific comparisons, and complete $n=10$--100 seed-42 curves for
\method{} and all three retrained steganalysis baselines. From $n=10$
to $n=100$, overall macro \tprone{} rises by 18.34\% and 16.10\% at resolution
256 for clean and watermarked negatives. At resolution 512, it rises by
26.13\% and 20.36\%. Increasing $n$ from 100 to 200 adds only 0.22--4.31\%. The
aggregate metrics improve through $n=100$ under fixed updates. Because the
update budget is fixed, these results do not define a learning curve with
matched per-example exposure, and individual cells need not be monotonic.

\begin{figure}[!htbp]
  \centering
  \includegraphics[width=\textwidth,height=0.38\textheight,keepaspectratio]{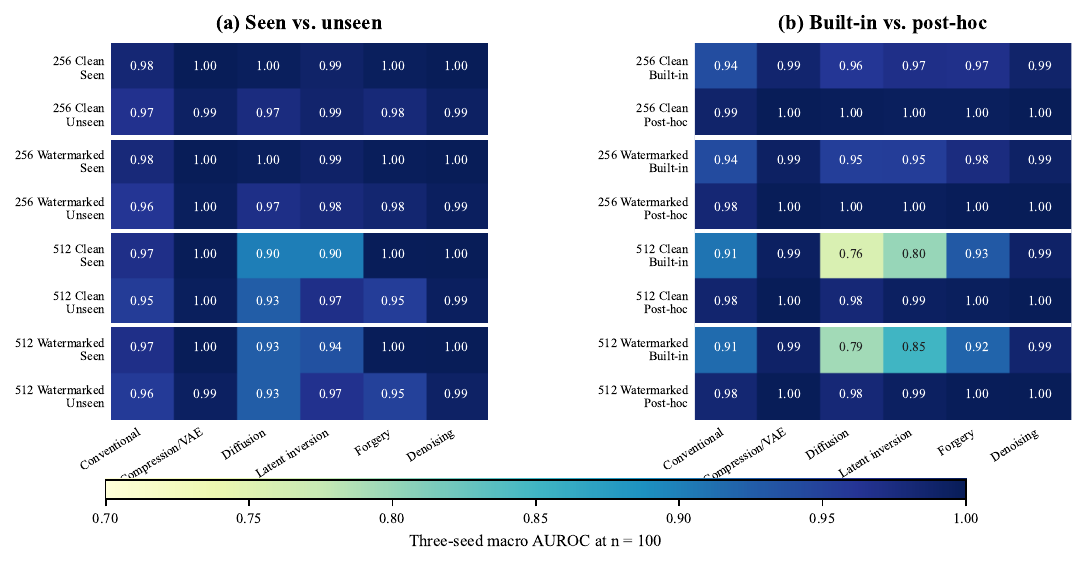}
  \caption{Attack-family detection with $n=100$ complete training triplets per
  attack. Cells report three-seed macro
  AUROC while keeping native resolution and negative role separate.
  Panel (a) compares seen and unseen watermarks; panel (b) compares built-in
  (generator-integrated) and post-hoc configurations. Family and group means
  macro-average independently computed metrics for each attack and watermark.}
  \label{fig:scaling}
\end{figure}

Figure~\ref{fig:scaling} reports heterogeneity in $n=100$ AUROC by attack
family. Panel (a) separates seen and unseen watermarks, whereas panel
(b) separates built-in and post-hoc embedding. Here, \emph{built-in} denotes
Stable Signature, Gaussian Shading, and Tree-Ring; \emph{post-hoc} denotes the
remaining seven configurations whose watermark is applied to an existing
image. At resolution 512, diffusion-regeneration AUROC is
75.69\%/79.44\% for built-in watermarks versus 98.06\%/98.10\% for post-hoc
watermarks under clean/watermarked negatives. The built-in
diffusion-regeneration subgroup at native resolution 512 has the lowest AUROC
in the figure. The
comparison does not attribute this difference to generator integration because
pairing regimes and group compositions differ.

\subsection{Comparison with retrained steganalysis systems}

Table~\ref{tab:baselines} reports the seed-42 overall comparison, and
Table~\ref{tab:family-baselines} gives the corresponding breakdown for six
attack families.
SiaStegNet has the highest \tprone{} among the baselines in all four overall
settings, where
\method{} improves \tprone{} by 10.80\%, 12.37\%, 13.87\%, and 15.68\%.
After macro-averaging both resolutions and both negative roles at the family
level, \method{} leads four columns, while SiaStegNet leads learned
compression/VAE reconstruction and classical denoising. Because all weights
are optimized for the present labels, these results compare task-adapted
systems, not off-the-shelf steganalysis checkpoints or loss-matched architecture
ablations. At seed 42, \method{} performs better
than the three task-adapted baselines in the four aggregate settings. The
comparison does not establish statistical superiority across seeds or
superiority over the broader image forensic literature.

\begin{table}[t]
\caption{Seed-42 \tprone{} (\%) with $n=100$ complete training triplets per
attack. Each entry macro-averages 23 attack
operations and ten watermarks for the specified resolution and negative role.
Every model is retrained from scratch on the same attack-specific forensic task.
Best values are bold.}
\label{tab:baselines}
\centering
\scriptsize
\begin{tabular}{@{}llrrrr@{}}
\toprule
Res. & Negative & SRNet & ZhuNet & SiaStegNet & \method{} \\
\midrule
256 & Clean       & 61.46 & 69.04 & 75.69 & \textbf{86.49} \\
256 & Watermarked & 63.33 & 68.34 & 75.40 & \textbf{87.77} \\
512 & Clean       & 51.61 & 61.80 & 70.25 & \textbf{84.12} \\
512 & Watermarked & 53.35 & 62.37 & 69.05 & \textbf{84.73} \\
\bottomrule
\end{tabular}
\end{table}

\begin{table}[!htbp]
\caption{Seed-42 family-level empirical test-ROC \tprone{} (\%) with $n=100$
complete training triplets per attack.
Entries macro-average slices defined by operation, watermark, resolution, and
negative role;
family operation counts appear in parentheses. Best values are bold.}
\label{tab:family-baselines}
\centering
\scriptsize
\setlength{\tabcolsep}{3pt}
\begin{tabular}{@{}lrrrrrr@{}}
\toprule
Model & \shortstack{Conventional\\distortions (12)} &
\shortstack{Learned compression/\\VAE reconstruction (5)} &
\shortstack{Diffusion\\regeneration (3)} &
\shortstack{Latent-space\\inversion (1)} &
\shortstack{Watermark\\forgery (1)} &
\shortstack{Classical\\denoising (1)} \\
\midrule
SRNet       & 52.12 & 93.61 & 19.94 & 27.25 & 44.77 & 95.75 \\
ZhuNet      & 60.84 & 91.44 & 46.08 & 21.12 & 63.65 & 93.75 \\
SiaStegNet  & 57.54 & \textbf{98.44} & 79.92 & 69.05 & 79.60 & \textbf{98.70} \\
\method{}   & \textbf{82.67} & 94.38 & \textbf{82.77} & \textbf{78.08} & \textbf{88.75} & 93.85 \\
\bottomrule
\end{tabular}
\end{table}

Attack-specific spectral and watermark-wise subgroup analyses are reported in
the supplement.

\subsection{Ablation study}
\label{sec:ablations}

\begin{figure}[!htbp]
  \centering
  \includegraphics[width=\textwidth]{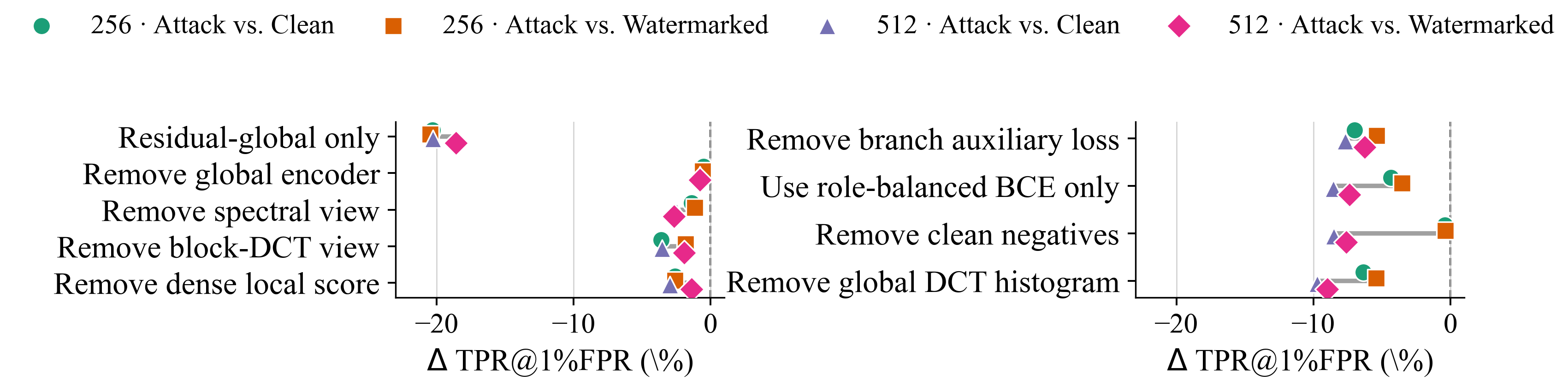}
  \caption{Effective ablations. Only settings that reduce \tprone{} in all
  four combinations of resolution and negative role are shown. Core-architecture
  and loss-suite points are three-seed means; clean-negative and global-DCT
  points use seed 42. Each point is the ablated setting minus its within-suite
  reference; negative values indicate degradation.}
  \label{fig:ablations}
\end{figure}

Figure~\ref{fig:ablations} reports only effective component removals. The
eight-operation core suite was fixed before result summarization to cover all
six attack families. We use a
prespecified directional criterion: the ablated setting must reduce \tprone{}
relative to its own reference in all four combinations of resolution and
negative role (256/512 $\times$ clean/watermarked). The suites differ in attack coverage and
seed count, so effect sizes remain within-suite comparisons rather than one
factorial ranking.

\paragraph{Effective components.}
On the eight-attack core suite, using only the residual-global path loses
18.58--20.47\%. Removing the global encoder or spectral view loses
0.50--0.79\% and 1.14--2.68\%, respectively. Removing the block-DCT view or
dense local score loses 1.85--3.62\% and 1.38--2.98\%, respectively. These
effects are measured on the eight-attack suite and need not hold uniformly
across all 23 pipelines. The supplement lists the eight operations.

Across all 23 attacks, removing branch auxiliary supervision loses
5.38--7.68\%, while role-balanced BCE alone loses 3.53--8.55\%. The full
objective has higher \tprone{} than either ablation on this suite. The BCE-only
comparison does not isolate the ranking, paired, gate, or HPF term.

At seed 42, removing clean negatives or the global DCT histogram loses
0.37--8.53\% and 5.43--9.73\%, respectively. Both rows meet the directional
criterion in all four conditions only at that seed. The supplement reports a
separate seed-42 table for all retained rows, so its values need not match the
multi-seed ranges above. Figure~\ref{fig:ablations} omits mixed-direction
settings.

\FloatBarrier

\section{Discussion and Limitations}

Attack-vs-Clean evaluates whether traces produced by watermark embedding and subsequent attack processing can be distinguished from paired clean-image controls. Attack-vs-Watermarked evaluates whether the additional attack processing can be distinguished from watermark presence alone. In copyright disputes, this task supplements direct watermark verification with evidence about processing history.

Existing watermarking systems typically verify image provenance through payload decoding, key matching, or detector scores \citep{zhu2018hidden,fernandez2023stablesignature,wen2023treering,yang2024gaussianshading}. When an attack compromises this evidence, the verification outcome alone cannot explain why verification failed. Attack forensics further examines whether an image contains processing traces consistent with a candidate watermark-removal process. If independent information points to a particular watermark-removal tool or algorithm, the corresponding specialist detector can assess whether a disputed image contains traces consistent with that pipeline. Its output can be interpreted together with the original files, watermark keys, digital signatures, and other investigative evidence to clarify the cause of verification failure and the processing history of the image. The detector alone, however, cannot establish copyright ownership, infringement, or attacker intent.

The current control protocol does not systematically include benign operations such as compression during transmission and platform re-encoding. Accordingly, the present experiments do not test whether malicious watermark removal can be distinguished from all types of benign post-processing. \method{} detects traces associated with a specified attack pipeline, but it cannot recover a damaged watermark payload or infer the key, owner identity, or signature information.

Future work will investigate whether estimates of the attack type and affected regions can guide image restoration and watermark decoding. This may allow the original ownership information to remain identifiable after an attack. Attack detection and payload recovery will therefore be combined so that, when direct verification fails, the forensic system can analyze the cause of failure and, where possible, recover evidence for ownership verification.

\section*{Acknowledgments}
This work was supported in part by Taishan Scholar under Grant tsqnz20250747;
in part by the National Natural Science Foundation under Grant 62502250,
Grant 62406051, Grant 62302249, Grant 62541206, and Grant 62272255; and in
part by the Young Talent of Lifting Engineering for Science and Technology in
Shandong under Grant SDAST2025QTB030.

\bibliography{iclr2027_conference}
\bibliographystyle{plainnat}

\end{document}